\documentclass[10pt,twocolumn,letterpaper]{article}

\usepackage{cvpr}              % To produce the CAMERA-READY version
\usepackage{graphicx}
\usepackage{amsmath}
\usepackage{amssymb}
\usepackage{booktabs}
\usepackage{multirow}
\usepackage{color}
\definecolor{mycitecolor}{rgb}{0, 0.4, 0.7}
\usepackage[breaklinks=true,bookmarks=true,citecolor=mycitecolor,colorlinks,pagebackref=true]{hyperref}
\usepackage[capitalize]{cleveref}
\crefname{section}{Sec.}{Secs.}
\Crefname{section}{Section}{Sections}
\Crefname{table}{Table}{Tables}
\crefname{table}{Tab.}{Tabs.}

\def\cvprPaperID{1691} % *** Enter the CVPR Paper ID here
\def\confName{CVPR}
\def\confYear{2023}

\begin{document}

\title{Stare at What You See: Masked Image Modeling without Reconstruction}

% \and

\author{Hongwei Xue$^{1,2}$\thanks{This work was performed when Hongwei Xue was visiting Shanghai AI Laboratory as a research intern.},~
Peng Gao$^{2,3}$\footnotemark[2],~
Hongyang Li$^2$\footnotemark[2],~
Yu Qiao$^2$,~
Hao Sun$^4$,~
Houqiang Li$^1$,~
Jiebo Luo$^5$\\
[2mm]
$^1$University of Science and Technology of China ~
$^2$Shanghai Artificial Intelligence Laboratory \\
$^3$ Shenzhen Institutes of Advanced Technology, Chinese Academy of Science \\
$^4$ China Telecom Corporation Ltd. Data\&AI Technology Company ~
$^5$University of Rochester
}

\maketitle
\renewcommand{\thefootnote}{\fnsymbol{footnote}}
\footnotetext[2]{Corresponding authors.}

\newcommand{\figintrotwo}{
\begin{figure}[t]
    \centering
    \includegraphics[width=\linewidth]{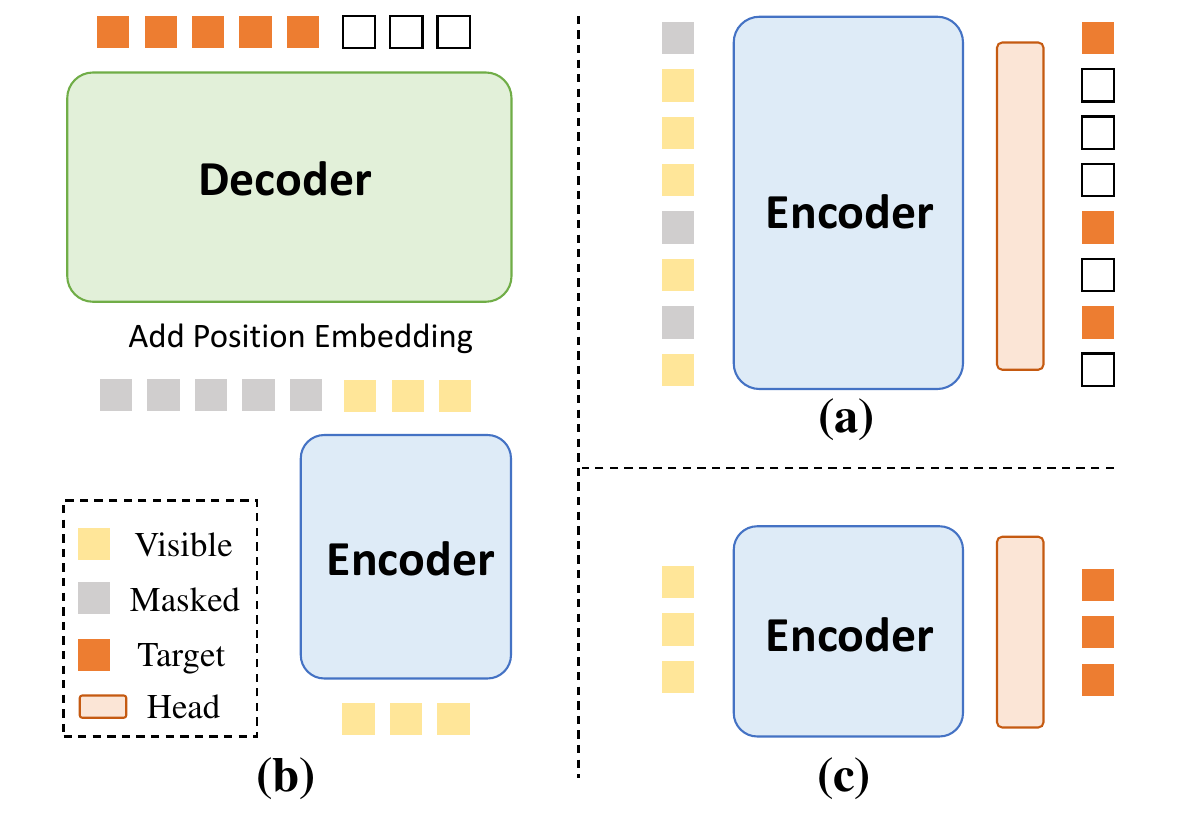}
    \vspace{-25pt}
    \caption{Comparison with existing paradigms of masked image modeling. \textbf{(a) Inpainting-style}: BEiT~\cite{bao2021beit}, MaskFeat~\cite{wei2022masked}, MVP~\cite{wei2022mvp}, BEiT V2~\cite{peng2022beit}, \textit{etc}. They take the whole image with some mask token replacement as the input of Encoder. Then a Linear head is applied to predict masked feature. \textbf{(b) Decoder-style}: MAE~\cite{he2022masked}, CAE~\cite{chen2022context}, MCMAE~\cite{gao2022convmae}, \textit{etc}. They drop most tokens and take the rest as the input of Encoder. Then a multi-layer Transformer is applied to decode masked features from visible tokens. \textbf{(c) Ours.}: Our paradigm take some visible tokens as the input of Encoder and align visible tokens with target features \textit{only}.}
    \vspace{-10pt}
    \label{fig:intro2}
\end{figure}
}

\newcommand{\figmodel}{
\begin{figure*}[t]
    \centering
    \includegraphics[width=0.8\linewidth]{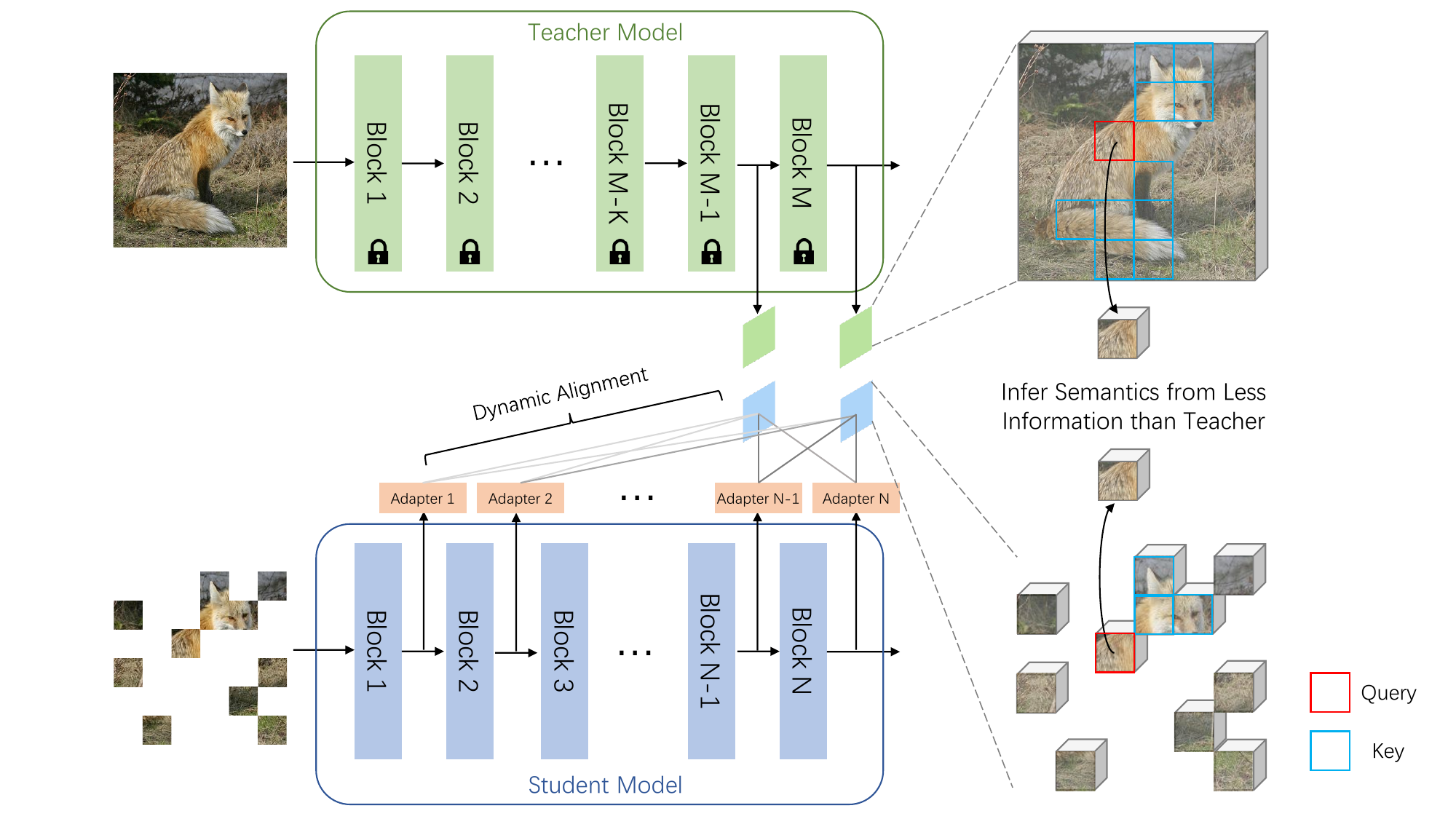}
    \caption{Framework of \textbf{MaskAlign}. MaskAlign aligns the visible features extracted by the student model and intact image features extracted by a frozen teacher model. Dynamic Alignment (DA) module learnably aggregates different level of student features and aligns with multi-level features of the teacher model.}
    \vspace{-10pt}
    \label{fig:model}
\end{figure*}
}

\newcommand{\figvis}{
\begin{figure*}[t]
    \centering
    \includegraphics[width=0.75\linewidth]{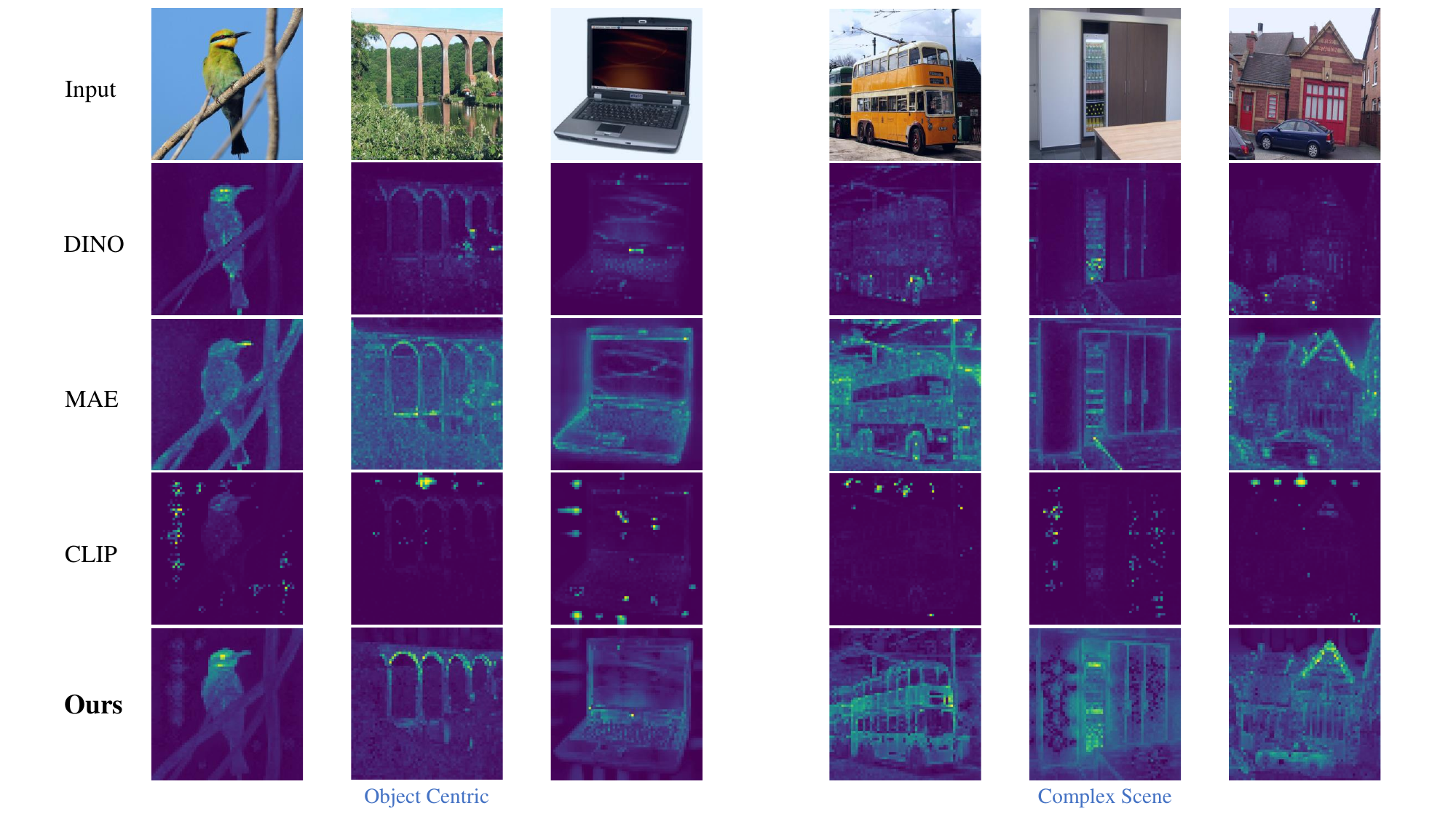}
    \caption{Visualization of attention weights at the last self-attention layer in DINO \cite{caron2021emerging}, MAE \cite{he2022masked}, CLIP \cite{radford2021learning} and ours. Object centric images are on the left while complex scenes are on the right. Our model highlights more reasonable regions.}
    \vspace{-10pt}
    \label{fig:vis}
\end{figure*}
}

\newcommand{\tabexpclass}{
\begin{table*}[t]
    \centering
    \begin{tabular}{l c c c c c c c} 
    \toprule
    Method & Backbone & Supervision & Forward Ratio & Reconstruct Ratio & Epochs & FT Acc.(\%)\\
    \midrule
    SimMIM~\cite{xie2022simmim} & Swin-B & RGB & 100\% & 60\% & 800 & 84.0  \\
    MCMAE~\cite{gao2022convmae} & CViT-B & RGB & 25\% & 75\% & 1600 & 85.0  \\
    MixMIM~\cite{liu2022mixmim} & MixMIM-B & RGB & 100\% & 100\% & 600 & 85.1 \\
    CMAE~\cite{huang2022contrastive} & CViT-B & RGB & 25\% & 75\% & 1600 & 85.3 \\
    \midrule
    BEiT~\cite{bao2021beit} & ViT-B & DALLE & 100\% & 40\% & 800 & 83.2  \\
    MAE~\cite{he2022masked} & ViT-B & RGB & 25\% & 75\% & 1600 & 83.6  \\
    CAE~\cite{chen2022context} & ViT-B & DALLE & 25\% & 75\% & 800 & 83.6  \\
    MaskFeat~\cite{wei2022masked} & ViT-B & HOG & 100\% & 40\% & 300 & 83.6  \\
    DMAE~\cite{bai2022masked} & ViT-B & MAE-L & 25\% & 75\% & 100 & 84.0  \\
    data2vec~\cite{baevski2022data2vec} & ViT-B & EMA & 100\% & 60\% & 800 & 84.2  \\
    MVP~\cite{wei2022mvp} & ViT-B & CLIP-B & 100\% & 40\% & 300 & 84.4  \\
    BEiT V2~\cite{peng2022beit} & ViT-B & CLIP-B & 100\% & 40\% & 300 & 85.0  \\
    MaskDistill~\cite{peng2022unified} & ViT-B & CLIP-B & 100\% & 40\% & 300 & 85.0  \\
    MILAN~\cite{hou2022milan} & ViT-B & CLIP-B & 25\% & 100\% & 400 & \bf85.4   \\
    \midrule
    FD-CLIP~\cite{wei2022contrastive} & ViT-B & CLIP-B & 100\% & \bf 0\% & 300 & 84.9   \\
    Ours & ViT-B & CLIP-B & 30\% & \bf 0\% & 200 & \bf85.4  \\
    \bottomrule
    \end{tabular}
    \caption{\textbf{Image classification} by fine-tuning on ImageNet-1K \cite{russakovsky2015imagenet}. 
    Our model achieves state-of-the-art performance with much fewer epochs.
    ‘Forward Ratio’ denotes the ratio of image tokens fed into the encoder. ‘Reconstruct Ratio’ denotes the ratio of reconstructed image tokens. ‘Epochs’ and ‘FT Acc.’ denote pre-training epochs and the top-1 accuracy of fine-tuning.}
    \vspace{-10pt}
    \label{tab:expclass}
\end{table*}
}

\newcommand{\tabablmask}{
\begin{table}[t]
    \centering
    \begin{tabular}{l c c c} 
    \toprule
    Model & Mask Type & Mask Ratio & FT Acc.(\%) \\
    \midrule
    \multirow{9}{*}{ViT-B/16} & \multirow{8}{*}{Attentive} & 0\% & 84.9 \\
     & & 20\% & 85.1 \\
     & & 40\% & 85.2 \\
     & & 60\% & \bf85.3 \\
     & & 70\% & \bf85.3 \\
     & & 75\% & 85.1 \\
     & & 80\% & 85.1 \\
     & & 90\% & 84.4 \\
    \cmidrule{2-4}
    & random & 70\% & 85.1 \\
    \bottomrule
    \end{tabular}
    \caption{Ablation study of mask ratio and strategy. MaskAlign performs well on wide range of mask ratios. We pre-train all models under the setting of Top 5 Dynamic Alignment with CLIP-B/16 teacher model.}
    \vspace{-10pt}
    \label{tab:ablmask}
\end{table}
}

\newcommand{\tabablalign}{
\begin{table}[t]
    \centering
    \begin{tabular}{l c c c} 
    \toprule
    Model & Align Type & Top $K$ & FT Acc.(\%) \\
    \midrule
    \multirow{6}{*}{ViT-B/16} & \multirow{4}{*}{Dynamic} & 1 & 84.9 \\
     & & 3 & 85.1 \\
     & & 5 & \bf85.3 \\
     & & 7 & 85.1 \\
     \cmidrule{2-4}
     & \multirow{2}{*}{Layer-wise} & 1 & 84.8 \\
     & & 5 & 85.1 \\
    \bottomrule
    \end{tabular}
    \caption{Ablation study of alignment strategy. MaskAlign achieves best results by Dynamic Alignment on Top 5 level feature. We pre-train all models on ImageNet for 200 epochs, under the setting of 70\% mask ratio with Attentive Masking strategy.}
    \vspace{-10pt}
    \label{tab:ablalign}
\end{table}
}

\newcommand{\tabablteacher}{
\begin{table}[t]
    \centering
    \begin{tabular}{l c c c} 
    \toprule
    Model & Teacher Model & T-FT Acc. & S-FT Acc. \\
    \midrule
    \multirow{3}{*}{ViT-B/16} & DINO-B & 82.8 \cite{caron2021emerging} & 83.9{\scriptsize ~(+1.1)} \\
    & CLIP-B & 82.9 \cite{wei2022contrastive} & 85.3{\scriptsize ~(+2.4)} \\
    & CLIP-L{\scriptsize 196} & - & 85.6 \\
    \midrule
    \multirow{2}{*}{ViT-L/16} & CLIP-B & 82.9 \cite{wei2022contrastive} & 86.5{\scriptsize ~(+3.6)} \\
     & CLIP-L{\scriptsize 196} & - & 87.4 \\
    \bottomrule
    \end{tabular}
    \caption{Comparison of different teacher model. T-FT and S-FT denote the finetuning Acc. of teacher model and student model. CLIP-L{\scriptsize 196} denotes input $196 \times 196$ resolution image to CLIP-L. We pre-train all models on ImageNet for 200 epochs.}
    \vspace{-10pt}
    \label{tab:ablteacher}
\end{table}
}

\newcommand{\tabablother}{
\begin{table}[t]
    \centering
    \begin{tabular}{l c c c c} 
    \toprule
    Model & Adaptor & Norm & [CLS] & FT Acc.(\%) \\
    \midrule
     \multirow{5}{*}{ViT-B/16} & Linear & LN & w/ & \bf85.4 \\
     & Linear & LN & w/o & 85.3 \\
     & MLP & LN & w/o  & 85.2  \\
     & Linear & BN & w/o &  84.7 \\
     & Linear & w/o  & w/o &  84.3 \\
    \bottomrule
    \end{tabular}
    \caption{Ablation study of other adaptation details. We pre-train all models on ImageNet for 200 epochs, under the setting of 70\% mask ratio with Attentive Masking strategy. Top 5 Dynamic Alignment with CLIP-B/16 teacher model.}
    \label{tab:ablother}
\end{table}
}

\newcommand{\tabspeed}{
\begin{table}[t]
    \centering
    \begin{tabular}{l c c c} 
    \toprule
    Model & Epochs & Time & FT Acc.(\%) \\
    \midrule
    MVP~\cite{wei2022mvp} & 300 & $2.8\times$ & 84.4  \\
    BEiT V2~\cite{peng2022beit} & 300 & $2.8\times$ &  85.0 \\
    MILAN~\cite{hou2022milan} & 400 & $3.2\times$ &  \bf85.4 \\
    FD-CLIP~\cite{wei2022contrastive} & 300 & $2.8\times$ & 84.9 \\
    Ours & 200 & $1.0\times$  &  \bf85.4 \\
    \bottomrule
    \end{tabular}
    \caption{Comparison of pre-training speed. MaskAlign achieves SOTA results with only about 1/3 training time. “Epochs” refer to the pre-training epochs of various methods. We compare with existing works with CLIP-ViT-B/16 as teacher model. Training speeds of these works are uniformed.}
    \label{tab:speed}
\end{table}
}

\newcommand{\tablinprobe}{
\begin{table}[h]
    \centering
    \begin{tabular}{l c c } 
    \toprule
    Model & Epochs & LP Acc.(\%) \\
    \midrule
    \textit{contrastive method} & \multicolumn{2}{c}{~} \\
    MoCov3 \cite{chen2021empirical} & 300 & 76.7 \\
    DINO \cite{caron2021emerging} & 400 & 78.2 \\
    \midrule
    \textit{MAE method} & \multicolumn{2}{c}{~} \\
    BEiT \cite{bao2021beit} & 800 & 56.7 \\
    SimMIM \cite{xie2022simmim} & 800 & 56.7 \\
    MAE \cite{he2022masked} & 1600 & 68.0 \\
    CAE \cite{chen2022context} & 800 & 68.3 \\
    MVP \cite{wei2022mvp} & 300 & 75.4 \\
    MILAN \cite{hou2022milan} & 400 & 78.9 \\
    BEiT v2 \cite{peng2022beit} & 300 & 80.1 \\
    FD-CLIP~\cite{wei2022contrastive} & 300 & 80.3 \\
    \midrule
    Ours & 200 & 79.9 \\ 
    \bottomrule
    \end{tabular}
    \caption{Comparison of the linear probing top-1 accuracy (LP Acc.) on ImageNet-1K dataset. “Epochs”
    refer to the pre-training epochs of various methods.}
    \label{tab:linprobe}
\end{table}
}

\newcommand{\tabexpcoco}{
\begin{table*}[t]
    \centering
    \begin{tabular}{l c c c c} 
    \toprule
    Method & Supervision & PT Epochs & AP{\scriptsize box} & AP{\scriptsize mask} \\
    \midrule
    Supervised \cite{he2017mask} & IN-1K Label & - & 47.9 & 42.9 \\
    MoCov3 \cite{chen2021empirical} & - & 300 & 47.9 & 42.7 \\
    DINO \cite{caron2021emerging} & - & 300 & 46.8 & 41.5 \\
    BEiT \cite{bao2021beit} & DALLE & 300 & 42.6 & 38.8 \\
    PeCo \cite{dong2021peco} & dVAE & 300 & 43.9 & 39.8 \\
    SplitMask \cite{el2021large} & dVAE & 300 & 46.8 & 42.1 \\
    CAE \cite{chen2022context} & DALLE & 800 & 49.2 & 43.3 \\
    MAE \cite{he2022masked} & RGB & 1600 & 50.3 & 44.9 \\
    ViTDet \cite{li2022exploring} & RGB & 1600 & 51.2 & 45.5 \\
    MILAN \cite{hou2022milan} & CLIP-B & 400 & \bf52.6 & 45.5 \\
    \midrule
    Ours & CLIP-B & 400 & 52.1 & \bf45.7 \\
    \bottomrule
    \end{tabular}
    \caption{\textbf{Object detection and instance segmentation results} obtained by finetuning Mask R-CNN \cite{he2017mask} on MS-COCO dataset \cite{lin2014microsoft}. All numbers are reported from the original paper and our model is implemented with the ViTDet protocol. All methods use ViT-B/16 pre-trained on ImageNet-1K dataset as the backbone. 
    Our model outperforms ViTDet \cite{li2022exploring} with shorten pre-training epochs from 1600 to 400.    
    “PT Epochs” refer to the pre-training epochs. }
    \vspace{-10pt}
    \label{tab:expcoco}
\end{table*}
}

\newcommand{\tabade}{
\begin{table}[ht]
    \centering
    \begin{tabular}{l c c } 
    \toprule
    Model & Epochs & mIoU(\%) \\
    \midrule
    \textit{contrastive method} & \multicolumn{2}{c}{~} \\
    MoCov3 \cite{chen2021empirical} & 300 & 47.3 \\
    DINO \cite{caron2021emerging} & 400 & 47.2 \\
    \midrule
    \textit{MAE method} & \multicolumn{2}{c}{~} \\
    BEiT \cite{bao2021beit} & 800 & 45.7 \\
    MAE \cite{he2022masked} & 1600 & 48.1 \\
    CAE \cite{chen2022context} & 1600 & 50.2 \\
    PeCo \cite{dong2021peco} &  300 & 46.7 \\
    MVP \cite{wei2022mvp} & 300 & 52.4 \\
    MILAN \cite{hou2022milan} & 400 & 52.7 \\
    \midrule
    Ours & 200 & 52.1 \\ 
    \bottomrule
    \end{tabular}
    \caption{Comparison of the semantic segmentation on ADE20K dataset. “Epochs”
    refer to the pre-training epochs of various methods.}
    \label{tab:seg}
\end{table}
}
\begin{abstract}
% modified a little bit by HY
Masked Autoencoders (MAE) have been prevailing paradigms for large-scale vision representation pre-training. By reconstructing masked image patches from a small portion of visible image regions, MAE forces the model to infer semantic correlation within an image. 
Recently, some approaches apply semantic-rich teacher models to extract image features as the reconstruction target, leading to better performance. 
However, unlike the low-level features such as pixel values, we argue the features extracted by powerful teacher models already encode rich semantic correlation across regions in an intact image. 
This raises one question: is reconstruction necessary in Masked Image Modeling (MIM) with a teacher model? In this paper, we propose an efficient MIM paradigm named MaskAlign. MaskAlign simply learns the consistency of visible patch features extracted by the student model and intact image features extracted by the teacher model.
To further advance the performance and tackle the problem of input inconsistency between the student and teacher model, we propose a Dynamic Alignment (DA) module to apply learnable alignment. Our experimental results demonstrate that masked modeling does not lose effectiveness even without reconstruction on masked regions. Combined with Dynamic Alignment, MaskAlign can achieve state-of-the-art performance with much higher efficiency.
% Our experimental results demonstrate that MaskAlign can both reduce the training computational cost and advance the performance, achieving state-of-the-art performances with much higher efficiency.
Code and models will be available at \url{https://github.com/OpenPerceptionX/maskalign}.

\end{abstract}

\section{Introduction}
\label{sec:intro}

\figintrotwo

In recent years, Vision Transformers are showing tremendous potential in computer vision area \cite{dosovitskiy2020image,touvron2021going,touvron2021training}. Following the big success of masked modeling in the natural language processing \cite{kenton2019bert}, Masked Image Modeling (MIM) has demonstrated a great ability of self-supervised learning \cite{he2022masked,bao2021beit}, while alleviating the data-hungry issue of Transformer architectures. The visual representation learned through MIM shows promising performance on various downstream vision tasks, outperforming the contrastive learning paradigms \cite{caron2021emerging,chen2021empirical}.

Existing Masked Image Modeling (MIM) methods aim to hallucinate the intact image from a small portion of visible image regions. As depicted in \cref{fig:intro2}, existing MIM methods are mainly divided into two types: (a) inpainting-style \cite{bao2021beit,xie2022simmim,wei2022masked} and (b) decoder-style \cite{he2022masked,chen2022context,gao2022convmae}. These two types both require the model to reconstruct masked regions. The inpainting-style models replace image regions with learnable vectors then fill them by the interaction within the encoder. The decoder-style models drop image regions then decode features from masked regions' positions based on the visible information. Some very recent works introduce semantic-rich teacher models like CLIP \cite{radford2021learning} into the two paradigms by using features extracted by teacher models as the reconstruction target \cite{wei2022mvp,hou2022milan,peng2022beit,peng2022unified}. In light of the semantic knowledge learned by teacher models, these works further improve the representation after masked image modeling, leading to better performance.

Reconstruction on masked regions implicitly forces the model's encoder to understand the semantic correlations within an image. However, the reconstruction manner brings much computation on masked tokens within or outside the encoder in inpainting-style or decoder-style, respectively. This redundant computation decreases the training efficiency of the encoder thus increasing the pre-training cost. Unlike low-level and isolated features such as normalized pixel values of patches, Histogram of Oriented Gradients (HOG), etc., the feature map extracted by powerful teacher models already contains rich semantic correlations, learned during the teacher model training stage. This difference raises one question: is reconstruction the only way in Masked Image Modeling (MIM) with teacher models? To answer this question, we propose a much more efficient MIM paradigm named \textbf{MaskAlign} without any reconstruction on masked tokens. 

On contrary of applying reconstruction on masked tokens, MaskAlign simply aligns the visible features extracted by the student model and intact image features extracted by the teacher model. As a consequence, MaskAlign forces the student model to learn not only good representation of the teacher model by feature alignment, but also the ability to hallucinate by masked modeling: feature consistency between the intact image and mask view requires the student model to infer semantics from much less information than teacher model. We adopt multi-level features of the teacher model as supervision to borrow richer semantics. However, the input of the student model contains much less information than the teacher model's, leading to misalignment of each layer's features. To tackle this problem, we enhance the student's features with a Dynamic Alignment (DA) module. DA dynamically aggregates different levels of student features and aligns with multi-level features of the teacher model. This approach can also easily transfer to asymmetric student-teacher structures. 

From our experimental results, MaskAlign with a wide range of mask ratio outperforms the mask ratio of 0\%, where it degenerates into Feature Distillation \cite{wei2022contrastive}. This verifies that masked modeling is still necessary for our paradigm. Meanwhile, our experiments validate the effectiveness of Dynamic Alignment by comparing different alignment strategies and numbers of feature levels. The combination of masked modeling and Dynamic Alignment makes our model achieve state-of-the-art results with much higher efficiency. For example, our model outperforms BEiT v2 \cite{peng2022beit} by 0.4\% on ImageNet Finetuning Accuracy (from 85.0\% to 85.4\%) with 1/3 pre-training time only. 

To sum up, our work has three-fold contributions:
\begin{enumerate}
    \item We categorize and rethink existing Masked Image Modeling (MIM) paradigms and propose 
    a more efficient MIM approach called MaskAlign. Even \textit{without} any reconstruction on masked tokens, MaskAlign achieves new state-of-the-art performance with much higher efficiency.
    \item We propose a Dynamic Alignment (DA) module to tackle the problem of input inconsistency between the student and teacher model, with negligible additional parameters and computation. 
    \item We conduct extensive experiments to verify the effectiveness of MaskAlign and Dynamic Alignment. Besides, our model shows a good ability of generalization on downstream tasks and larger size models. 
\end{enumerate}

\section{Related Work}
\figmodel
\paragraph{Masked Image Modeling.}
Motivated by Masked Language Modeling (MLM) in BERT \cite{kenton2019bert}, BEiT \cite{bao2021beit} explores Masked Image Modeling (MIM) on vision transformers by reconstructing the dVAE \cite{rolfe2016discrete} feature extracted by DALL-E \cite{ramesh2021zero}. MAE \cite{he2022masked} and SimMIM \cite{xie2022simmim} find that RGB values can act as a simple yet good enough reconstruction target for masked modeling. PeCo \cite{dong2021peco},  iBOT \cite{zhou2021image} and MaskFeat \cite{chen2021empirical} respectively use dVAE with perceptual loss, an online tokenizer and the manually-crafted HOG descriptor, proving that the reconstruction target shows big impact. Motivated by this, MVP \cite{wei2022mvp} firstly introduces multimodality-guided teacher models into MIM, by simply replacing the reconstruction target with CLIP \cite{radford2021learning} features. Rich-semantic guidance leads to impressive gains. Some very recent works: BEiT V2 \cite{peng2022beit}, MILAN \cite{hou2022milan} and MaskDistill \cite{peng2022unified} also include CLIP in their model. BEiT V2 
% uses
adopts
CLIP features to train their discrete tokenizer, while MILAN and MaskDistill directly use CLIP features as reconstruction targets. All existing MIM works are based on reconstruction. In this paper, we explore a new paradigm of MIM without reconstruction, 
% needed, 
which significantly alleviates the efficiency issue brought by redundant computation of reconstruction.

\paragraph{Vision Language Pre-training.}
Learning visual linguistic representations from large-scale data has demonstrated unprecedented power in cross-modal learning \cite{radford2021learning,jia2021scaling,zellers2021merlot,xue2022advancing,xue2022clip,sun2022long,fu2021violet,hu2022scaling,geng2023hiclip}. Under the guidance of diverse texts with rich semantics, CLIP \cite{radford2021learning} advances the transfer performance on many downstream vision tasks, especially image-text generation \cite{patashnik2021styleclip,mokady2021clipcap}, requiring a sufficient understanding of semantic correlations within an image. Some existing masked modeling methods adopt the CLIP feature as the reconstruction target \cite{wei2022mvp,peng2022beit,hou2022milan}, 
outperforming counterparts using low-level features \cite{he2022masked,chen2021empirical}. 
In this work, 
we also adopt a frozen CLIP model to leverage good semantics and further improve the representation ability by incorporating masked image modeling.

\paragraph{Knowledge Distillation.}
Knowledge distillation (KD) \cite{hinton2015distilling} generates a soft label by the output of the teacher model to train the student model. KD transfers the capacity of teacher models into students and brings impressive gains. From that, KD has shown great potential in various tasks \cite{touvron2021training,he2019knowledge,yang2021knowledge} and domains \cite{jiao2020tinybert,wang2020minilm}. Feature Distillation (FD) \cite{wei2022contrastive} finds that using the normalized dense feature from the teacher model to supervise student models can significantly advance the performances. In this paper, we leverage mask modeling in aligning with the teacher model instead of full-size input, leading to significant improvement in both performance and training efficiency.
\section{Approach}
MaskAlign aligns the visible features extracted by the student model and intact image features
extracted by a frozen teacher model. The overview of MaskAlign is depicted in \cref{fig:model}. In this section, we elaborate on the details of masking and alignment.

\subsection{Model Structure}
MaskAlign consists of a randomly initialized student model and a pre-trained teacher model. For the student model, we adopt standard Vision Transformer (ViT) as in \cite{dosovitskiy2020image} to make a fair comparison with existing works. We apply a frozen teacher model with rich semantics to produce the supervision. In experiments, we adopt ViT teacher models such as CLIP-ViT \cite{radford2021learning} and DINO \cite{caron2021emerging}. For the input of teacher models, an image $I \in \mathbb{R}^{C \times H \times W}$ is divided to $N = HW/P^2$ patches: $\mathcal{I} = \left\{\boldsymbol{x}_i^p\right\}_{i=1}^N$ and $\mathcal{I} \in \mathbb{R}^{N \times\left(P^2 C\right)}$, where $(P, P)$ is the patch size. The image patches $\mathcal{I}$ are then linearly projected to input tokens with added positional embeddings. For the input of student models, the process is similar except we use a masked view of the image. Like in MAE, we drop $r\%$ patches and only feed visible patches $\mathcal{V} = \left\{\boldsymbol{x}_i^p\right\}_{i=1}^{N(1-r\%)}$ into student model. By Self-Attention mechanism within the Transformer, patches interact with others to aggregate information. In the student model, each patch (query) can only attend to $N(1-r\%)$ patches (keys), which are much less than the teacher model. This can greatly decrease the training cost and encourage the student model to learn a better ability of visual representation.

\subsection{Masking Strategy}
To eliminate the redundancy of an image, masking creates a task that cannot be easily solved by extrapolation from visible neighboring patches. To generate a mask view $\mathcal{V}$ from an intact image $\mathcal{I}$, one straightforward sampling strategy is random masking, which samples patches without replacement, following a uniform distribution \cite{he2022masked}. 

Another masking strategy is based on the guidance of the teacher model. Following  \cite{xue2021probing,zellers2021merlot,kakogeorgiou2022hide,hou2022milan}, we also study attentive masking in our paradigm. Attentive masking aims to feed tokens covering important image regions into the encoder with high probabilities. By doing so, the latent representations from the encoder provide sufficient information to infer semantics. We make a comparison of these two masking strategies in \cref{sec:experiment}.

\subsection{Dynamic Alignment}
To borrow richer semantics from the teacher model, we use multi-level features as supervision. However, aligning student model's features with multi-level supervision has a challenge: as the input of the student model contains much less information than the teacher’s, the input inconsistency causes misalignment between the student and teacher model on each layer. To tackle this problem, we propose a Dynamic Alignment (DA) module. DA can dynamically learn how to make alignment between the student and teacher model. A Transformer usually consists of a sequence of blocks. We add one adaptor $A_i$ to each block's output $x_i$ to project the student model's feature space to the teacher's. The adaptor could be a light model like a Linear layer or 2-layer MLP. To dynamically aggregate different levels of student features and align with multi-level features of the teacher model, we apply a Dynamic Alignment Matrix: $W$, which is an $S \times T$ matrix with entries $w_{ij}$, where $S$ and $T$ is the number of blocks in student and teacher model. The whole DA module can be formulated as:
\begin{equation}
    \hat{y} = \{\sum_{i=0}^{S} w_{ij}A_i(x_i)\}_{j=0}^T.
\end{equation}
where $\hat{y}$ is a set of linear combinations of multi-level features of the student model. During pre-training, gradients can backpropagate to Dynamic Alignment Matrix. In downstream tasks, we simply abandon the whole DA module.

To restrain the feature magnitudes of teacher features, we generate the alignment target $\tilde{y}$ by normalizing each level of teacher features as MAE \cite{he2022masked} does on pixel values: 
\begin{equation}
    \tilde{y} = \{\text{Normalize}(y_i)\}_{j=0}^T.
\end{equation}

Finally, following \cite{wei2022contrastive}, we employ a smooth L1 loss between the student and teacher features:
\begin{equation}
\mathcal{L}_{\text {Align }}(\hat{y}, \tilde{y})= \begin{cases}\frac{1}{2}(\hat{y}-\tilde{y})^2 , 
& \mid \hat{y}-\tilde{y} \mid \leq 1 \\ 
\left(\mid \hat{y}-\tilde{y} \mid-\frac{1}{2} \right), & \text { otherwise }\end{cases}.
\end{equation}
In experiments, we align student features with the latter layers of the teacher model, determined by a hyperparameter $K$. We also compare with alignment without a Dynamic Alignment Matrix, namely Layer-wise alignment. Layer-wise denotes aligning features layer-by-layer without a Dynamic Alignment Matrix. Experimental results in \cref{sec:experiment} demonstrate that Dynamic Alignment outperforms simple Layer-wise alignment, with nearly no increase in computational effort.

\subsection{Relation to Existing Models}
As depicted in \cref{fig:intro2}, in this work we explore a new paradigm for masked image modeling. Inpainting-style models like BEiT V1/V2 \cite{bao2021beit,peng2022beit}, MaskFeat \cite{chen2021empirical}, MVP \cite{wei2022mvp} simultaneously process the partially masked image content and produce predictions for the masked patches. The full-size input leads to large computational costs for this kind of model. Decoder-style models like MAE \cite{he2022masked}, CAE \cite{chen2022context} and MCMAE \cite{gao2022convmae} only take the partial image as the input. The encoder maps the input into a latent representation, and the decoder reconstructs the input from the latent representation. The decoder learns the interaction on full-size features but is abandoned after pre-training. Both of these paradigms have the redundant computation of reconstruction. On the contrary of them, our model MaskAlign does not include any reconstruction on masked tokens. Our model only applies alignment on visible features extracted by the student model and intact image features extracted by the teacher model. As a result, our model has big advantages in efficiency and simplicity.
\section{Experiments}\label{sec:experiment}

\tabexpclass
\subsection{Implementation}
\paragraph{Pre-training.} 
We pre-train MaskAlign on ImageNet-1k dataset \cite{russakovsky2015imagenet}, containing 1.28M training images. For the student model, we mainly study ViT-B/16 (12 blocks, 768 hidden size) and ViT-L/16 (24 blocks, 1024 hidden size). We employ the default $16 \times 16$ input patch size, partitioning the image of $224 \times 224$ into $14 \times 14$ patches. We only use standard random cropping and horizontal flipping for data augmentation, following MAE \cite{he2022masked}, CAE \cite{chen2022context}, etc. For pre-training, we train all models using AdamW optimizer \cite{kingma2014adam}, with a base learning rate of 1.5e-4, weight decay of 0.05, and optimizer momentum $\beta_1$, $\beta_2$ = 0.9, 0.95. We use a total batch size of 1024, and pre-train models under a cosine learning rate decay schedule with $10\%$ warm-up steps. We also employ stochastic depth \cite{huang2016deep} with a 0.1 rate, and disable dropout in the Linear layer and Self-Attention. 
\paragraph{Image Classification.} 
After pre-training, we evaluate the model by fine-tuning on ImageNet-1K and report the top-1 accuracy on the validation set. We fine-tune our model 100 epochs with 5 warm-up epochs. We use the same batch size,
optimizer, and weight decay as in pre-training. The base learning rate, layer-wise learning rate decay,
and drop path rate are set to be 3e-4, 0.6 and 0.2, respectively. We use the Data Augmentation in MAE without any modification.

\paragraph{COCO Detection and Instance Segmentation.}
We also evaluate on COCO dataset \cite{lin2014microsoft} for object detection and instance segmentation to verify the transferability of our model. We follow the benchmark ViTDet \cite{li2021benchmarking,li2022exploring}, the pre-trained backbone is adapted to FPN \cite{lin2017feature} in the Mask R-CNN framework \cite{he2017mask}. The resolution of the input image, learning rate, and layer decay are respectively set as $1024 \times 1024$, 3e-4 and 0.8. The model is fine-tuned for 25 epochs with a total batch size of 64. As finetuning all methods is heavy and has the potential risk of non-optimal results, we take other results from original papers.

\subsection{Comparison to State-of-the-arts}
\paragraph{Classification.}
\cref{tab:expclass} shows the comparison of ImageNet finetuning results between our model and previous state-of-the-art approaches of the similar model size. We also list the Forward Ratio and Reconstruction Ratio in \cref{tab:expclass} to intuitively compare the role of encoder in each paradigm. For example, the encoders of BEiT V1/V2 \cite{bao2021beit,peng2022beit}, MaskFeat \cite{chen2021empirical} and MVP \cite{wei2022mvp} process the full-size input with 40\% masked token. During the encoding, these 40\% masked tokens will be filled under the supervision of reconstruction targets. This kind of paradigm includes invalid information in inputs, leading to efficiency damage and risks of a gap between pre-training downstream tasks. For MAE \cite{he2022masked} and MILAN \cite{hou2022milan}, the encoder and decoder respectively process 25\% and 100\% patches. This paradigm still suffers from the computation on the decoder as it is totally abandoned in downstream tasks. A recent work FD-CLIP \cite{wei2022contrastive} uses the normalized feature of CLIP for distillation, thus the encoder process the full-size input. Compared to them, our model only processes 30\% patches. While significantly reducing the training cost (1/3 Forward Ratio and 2/3 PT Epochs), our model outperforms BEiT V2, MaskDistill and FD-CLIP on Top-1 Accuracy (85.4\% vs. 85.0\% and 84.9\%).  

\paragraph{Detection and Segmentation.}
To verify the generalization of our method, we evaluate on COCO object detection and instance segmentation, by adapting the pre-trained ViT-B/16 backbone to FPN \cite{lin2017feature} in the Mask R-CNN framework \cite{he2017mask}. The results are shown in \cref{tab:expcoco}. Our model achieves 52.1\% on AP{\scriptsize box} and 45.7\% on AP{\scriptsize mask}. Our model outperforms ViTDet \cite{li2022exploring} with shortened pre-training epochs from 1600 to 400 and fine-tuning epochs from 100 to 25.

\tabexpcoco
\subsection{Ablation Study}
\tabablalign
\paragraph{Dynamic Alignment.}
To verify the effectiveness of the Dynamic Alignment (DA) module, we conduct a series of experiments of making comparisons between w/ and w/o DA and different top $K$s. The results are shown in \cref{tab:ablalign}. To fairly compare, We pre-train all models on ImageNet for 200 epochs, under the setting of 70\% mask ratio with Attentive Masking strategy and CLIP-B/16 as the teacher model. For alignment type, Dynamic denotes using our proposed DA module and Layer-wise denotes aligning features layer-by-layer without a Dynamic Alignment Matrix. From \cref{tab:ablalign}, Multi-level alignment target performs better than only aligning with the teacher model's final output (85.3\% vs. 84.9\%). 
Multi-level features from the teacher model can provide richer semantics as supervision. It's intuitive that the latter blocks of the teacher model provide more high-level than former ones. As a result, a proper $k$ has a big impact. We search the best of 3, 5, 7 for the hyper-parameter top $K$, and keep it fixed for other experiments in this paper. Compared with Layer-wise type, Dynamic type achieves better performance (85.3\% vs. 85.1\%). This validates the effectiveness of our Dynamic Alignment (DA) module.
It's worth noting that Layer-wise type with top 1 is equivalent to feature distillation on visible tokens, and our method gains 0.5\% improvement on it. These results verify that the combination of MaskAlign and Dynamic Alignment will significantly improve the pre-training.

\paragraph{Mask Strategy.}
As our paradigm is totally different from existing paradigms based on reconstruction, the impacts of mask ratio could be also different. In this part, we compare different mask ratios and strategies. As MaskAlign only uses $1-r\%$ features produced by the teacher model per iteration, different mask ratios will result in unmatched training efficiency. To make a fair comparison, we adjust the training iterations for each experiment to be equal to 200 epochs for 70\% mask ratio. We pre-train ViT-B/16 as the student model under the setting of Top 5 Dynamic Alignment with CLIP-B/16 as the teacher model. The results are shown in \cref{tab:ablmask}. From \cref{tab:ablmask}, interestingly, we observe that although the peak accuracy is gained at around 60-70\%, the performance and mask ratio curve is much more flat than MAE's \cite{he2022masked}. For example, when mask ratio change from 60\% to 20\%, the FT Acc. of MAE drops 1.6\%  while ours only drops 0.2\%. And the curve is also flat at high mask ratios. For reconstruction, excessive visible patches will lead to a simplistic pre-training task while insufficient visible patches will fail in providing necessary information for the model to infer masked patches. As a result, reconstruction-based methods need more careful choice of mask ratios. For alignment on the visible features, even a 0\% mask ratio gains improvement compared with the teacher model \cite{wei2022contrastive}. Introducing masked modeling into distillation will further improve the performance, yet relies less on mask ratio.

\tabablmask
\paragraph{Teacher Model.}
To study the ability of generalization and scaling-up behavior, we also conduct experiments on comparing impacts by different teacher models. In this part, we pre-train each model for 200 epochs, under the setting of Top 5 Dynamic Alignment and 70\% mask ratio. We choose DINO-B/16 \cite{caron2021emerging}, CLIP-B/16 and CLIP-L/14 as teacher models to train ViT-B/16. To match the student model's feature size, we resize the input resolution of CLIP-L/14 to $196 \times 196$. To study the ability of generalization on larger models, we choose CLIP-L/14 as the teacher model to train a ViT-L/16 model. The results are shown in \cref{tab:ablteacher}. We find that MaskAlign consistently works well on various teacher models. Student models outperform teacher models in the same size: for base size models, MaskAlign leads to 1.1\% and 2.4\% improvement on DINO and CLIP, respectively. For a large-size model ViT-L/16, MaskAlign leads to 3.6\% improvement on CLIP-B/16. These results validate that our method has a good ability to generalize on different teacher models.
\tabablteacher
\paragraph{Adaptation Details.}
We also compare different Adaptors and target normalization methods. For Adaptor, we compare a simple Linear layer with a 2-layer MLP, with the same hidden size as the dimension of features from the student model. For normalization, we compare LayerNorm, BatchNorm and original feature. We disable learnable scale and bias for both LayerNorm and BatchNorm. We pre-train all models on ImageNet-1K for 200 epochs, under the setting of 70\% mask ratio with Attentive Masking strategy. Top 5 Dynamic Alignment with CLIP-B/16 teacher model. From the results in \cref{tab:ablother}, MLP adaptor does not bring performance improvement compared to a simple Linear Layer, even though more parameters are included in alignment. 
Our results show that the type of normalization has a big impact, LayerNorm performs much better than BatchNorm and no normalization. This observation is also consistent with \cite{wei2022contrastive}.
We find that including [CLS] token in alignment will slightly improve the accuracy. As [CLS] of CLIP aggregates global information of the image, mimicking [CLS] helps MaskAlign learn more knowledge of the interaction between different patches. Results in Tab. \ref{tab:ablalign}, \ref{tab:ablmask} and \ref{tab:ablteacher} are under w/o [CLS] token setting.

\tabablother
\tabspeed

\figvis
\paragraph{Pre-training Speed.}
We compare the pre-training speed of our model with state-of-the-art methods. MVP \cite{wei2022mvp}, BEiT V2 \cite{peng2022beit} and MILAN \cite{hou2022milan} are all Masked Image Modeling pre-training models using CLIP-ViT-B/16 features as reconstruction target. Specifically, MVP and BEiT V2 are in inpainting-style. They mask input patches by replacing with learnable tokens then filling these patches, thus the encoder processes a full-size input. MILAN adopts decoder-style, the encoder only processes visible patches and full-size intermediate features are handled by a decoder. Although the decoder is much lighter than the encoder, its full-size input still brings much computation. Besides, the decoder only plays a role in pre-training stage, and is abandoned in downstream tasks. This gap limits the efficiency of the pre-training. A recent work FD-CLIP \cite{wei2022contrastive} distills the feature map of CLIP model at the output. Compared with them, our model has the lightest architecture in theory. 

\cref{tab:speed} lists the pre-training times and ImageNet-1K finetuning performances. The training times are uniformed. Our model outperforms MVP, BEiT V2 and FD-CLIP, and achieves comparable performances with MILAN, but with only about 1/3 training time. It's worth noting that our model only uses $1-r\%$ features produced by the teacher model per iteration. As a result, the inference time of the teacher model makes our method not have a linear acceleration ratio. Our method's acceleration is more significant in scenarios when teacher model is lighter than the student model, e.g., CLIP-B/16 supervises ViT-L/16.

\paragraph{Attention Visualization.}
To intuitively peek at what is learned in MaskAlign during pre-training, we visualize the attention map of [CLS] token of the last self-attention layer of different models. The comparison is shown in \cref{fig:vis}. DINO is trained by contrastive learning, which minimizes the similarities between augmented views of images. Random cropping in view augmentation makes DINO tend to focus mainly on the salient region in the original image. Thus the attention weights of DINO are usually concentrated on one salient object. MAE reconstructs masked pixels from visible regions, thus more texture information is learned, leading to a waste of capacity on low-level features irrelevant for semantic understanding. CLIP has good semantic alignment with language, however, we surprisingly find that CLIP features have bad correspondence to semantic regions. This may caused by its sparse supervision of texts. Although supervised by CLIP, our model's attention map seems more reasonable on both object-centric images and complex scenes. MaskAlign accurately concentrates on salient objects. While handling complex scenes, MaskAlign covers different semantic regions in one image.

\section{Conclusion}

In this paper we first categorize and rethink existing Masked Image Modeling (MIM) paradigms. Both inpainting-style and decoder-style models need much computation on masked tokens, decreasing the training efficiency of pre-training. Following some approaches that apply semantic-rich teacher models to extract image features as supervision, we propose a MIM paradigm named MaskAlign without any reconstruction. MaskAlign simply aligns the visible features extracted by the student model and intact image features extracted by the teacher model. And we propose a Dynamic Alignment (DA) module to tackle the problem of input inconsistency between the student and teacher model. We conduct extensive experiments to verify the effectiveness of our method. Our model achieves state-of-the-art performances with much higher pre-training efficiency. In the future, we will explore the scaling up of MaskAlign for vision recognition.

\section{Broader Impact}
Our work explores a new paradigm for masked image modeling, which may encourage future works to reconsider the role of masking in pre-training or distillation. On contrary of existing models based on reconstruction, MaskAlign borrows the teacher model's semantic information to learn feature consistency and without any supervision on masked tokens, our method demonstrates a strong ability of representation and efficiency. Besides, MaskAlign has an extremely light and simple framework. In the future, we may move forward to 1) find more mathematical explanations of MaskAlign, and 2) transfer MaskAlign to large-scale multi-modal pre-training to leverage its advantages in both efficiency and simplicity

\paragraph{Acknowledgement} This work is partially supported by the National Natural Science Foundation of China (Grant No.62206272), National Key R\&D Program of China (NO.2022ZD0160100), and in part by Shanghai Committee of Science and Technology (Grant No. 21DZ1100100).

%%%%%%%%% REFERENCES
{\small
\bibliographystyle{ieee_fullname}
\bibliography{egbib}
}

\clearpage
\appendix
\section{Linear Probing}
We also perform linear probing by appending a linear classifier after the final layer of the pre-trained model, following MAE \cite{he2022masked}. We use a mini-batch size of 16384, and an initial base learning rate of 0.025 for ViT-Base. We train the linear classifier for 90 epochs. The learning rate is linearly warmed up for the first 10 epochs, and decayed to zero by a cosine learning rate schedule. We do not use 
mixup, cutmix, drop path, or color jittering, and set the weight decay to zero. For other methods, we report the number from original papers. 

\tablinprobe

From the results, MaskAlign outperforms methods based on contrastive learning, such as MoCov3 and DINO, and masked modeling methods based on CLIP, such as MVP and MILAN.

\section{ADE20K Semantic Segmentation}
We transfer our pre-trained backbone models to semantic segmentation task on the ADE20K dataset \cite{zhou2017scene}. Following MAE, the ViT models pre-trained on ImageNet-1K dataset serve as the backbone of UperNet \cite{xiao2018unified}, and are finetuned together with the segmentation layers. We report the mean intersection over union (mIoU) averaged over all semantic categories. 

\tabade

From the results, MaskAlign achieves comparable segmentation performance with MVP, MILAN, but with much fewer pre-training epochs.

\section{ImageNet-9 Backgrounds Challenge}
We add the ImageNet-9 \cite{xiao2020noise} Backgrounds challenge under linear probing setting. We follow the protocol of AttMask \cite{kakogeorgiou2022hide} and report results on: Only-FG (OF), Mixed-Same (MS), Mixed-Rand (MR), and Mixed-Next (MN), No-FG (NF), and original. 

\begin{table}[h]
    \centering
    \small
    \begin{tabular}{l c c c c c c} 
    \toprule
    Model & OF & MS & MR & MN & NF & Original \\
    \midrule
    AttMask \cite{kakogeorgiou2022hide} & 75.2 & 76.2 & 62.3 & 59.4 & 40.6 & 89.8 \\
    MAE~\cite{he2022masked} & 81.3 & 77.8 & 66.3 & 64.0 & 38.6 & 91.9 \\
    MILAN~\cite{hou2022milan} & \bf89.2 & 87.1 & 77.9 & 74.7 & 46.8 & 96.2 \\
    Ours & 87.7 & \bf87.2 & \bf78.6 & \bf76.6 & \bf51.9 & \bf96.4 \\
    \bottomrule
    \end{tabular}
    \caption{Comparison of the linear probing results on ImageNet-9 dataset. }
    \label{tab:reb_in9}
\end{table}

MaskAlign has better robustness than MILAN. As Background has a higher probability of being masked and reconstructed (due to SAS in MILAN), the removal of reconstruction makes the model focus more on the foreground.

\end{document}